\documentclass{article}

\usepackage{microtype}
\usepackage{graphicx}
\usepackage{subcaption}
\usepackage{booktabs} 

\usepackage{hyperref}

\usepackage[preprint]{icml2026}

\usepackage{amsmath}
\usepackage{amssymb}
\usepackage{mathtools}
\usepackage{amsthm}
\usepackage{multirow}
\usepackage{graphicx} 
\usepackage{lipsum}   

\usepackage[capitalize,noabbrev]{cleveref}

\theoremstyle{plain}

\theoremstyle{definition}

\theoremstyle{remark}

\usepackage[textsize=tiny]{todonotes}

\icmltitlerunning{IDO: Incongruity-aware Distribution Optimization for Multimodal Fake News Detection}

\begin{document}

\twocolumn[
  \icmltitle{IDO: Incongruity-aware Distribution Optimization for \\ Multimodal Fake News Detection}


  \icmlsetsymbol{equal}{*}

  \begin{icmlauthorlist}
    \icmlauthor{Hengyang Zhou}{equal,yyy}
    \icmlauthor{Rongman Hong}{equal,comp}
    \icmlauthor{Yuxuan Zhou}{equal,yyy}
    \icmlauthor{Jing Wang}{ecnu}
    \icmlauthor{Zhaoyan Pan}{zju}
  \end{icmlauthorlist}

  \icmlaffiliation{yyy}{Nanjing University}
  \icmlaffiliation{comp}{University of Birmingham}
  \icmlaffiliation{ecnu}{East China Normal University}
  \icmlaffiliation{zju}{Zhejiang University}

  \icmlcorrespondingauthor{Hengyang Zhou}{hyzhou03@gmail.com}

  \icmlkeywords{Machine Learning, ICML}

  \vskip 0.3in
]



\printAffiliationsAndNotice{\icmlEqualContribution}

\begin{abstract}
Multimodal fake news detection aims to identify the authenticity of news. Existing multimodal fake news detection methods mainly focus on cross-modal consistency, but often fail to explicitly model the semantic incongruity that characterizes deceptive multimodal content. However, misinformation often contains semantic information incongruity with the facts. To address these challenges, we propose  Incongruity-aware Distribution Optimization (IDO) to improve the performance of fake news detection from the perspectives of factual incongruity and modality incongruity. For factual incongruity, we introduce a channel-wise reweighting strategy to obtain semantically discriminative embeddings and utilize gaussian distribution to model the uncertain correlation caused by  factual incongruity. For modality incongruity, we utilize incongruity contrastive learning to learn cross-modal semantic information. Experiments demonstrate that IDO achieves state-of-the-art performance.
\end{abstract}

\section{Introduction}
Multimodal fake news detection (MFND) poses significant challenges, as deceptive information is often jointly expressed across multiple modalities (e.g., textual and visual content). Multimodal large language models possess powerful generative capabilities, which have significantly reduced the cost of manufacturing fake news. Additionally, they can generate and spread misinformation due to AI hallucination \citep{goldstein2023generative,nslm}. Multimodal forgery media, by fusing and spreading a wider range of information, has greater misleading potential on social media \citep{fka-owl}. On the other hand, the internet is increasingly filled with multimodal tweets (e.g., text and image-based), which are well-known for their high attractiveness and deceptive properties \citep{cao2020exploring}. Therefore, developing automated detection systems to curb the spread of multimodal fake news has become an urgent necessity.

\begin{figure}[!t]
    \centering
    \begin{subfigure}{0.45\columnwidth}
        \includegraphics[width=\textwidth]{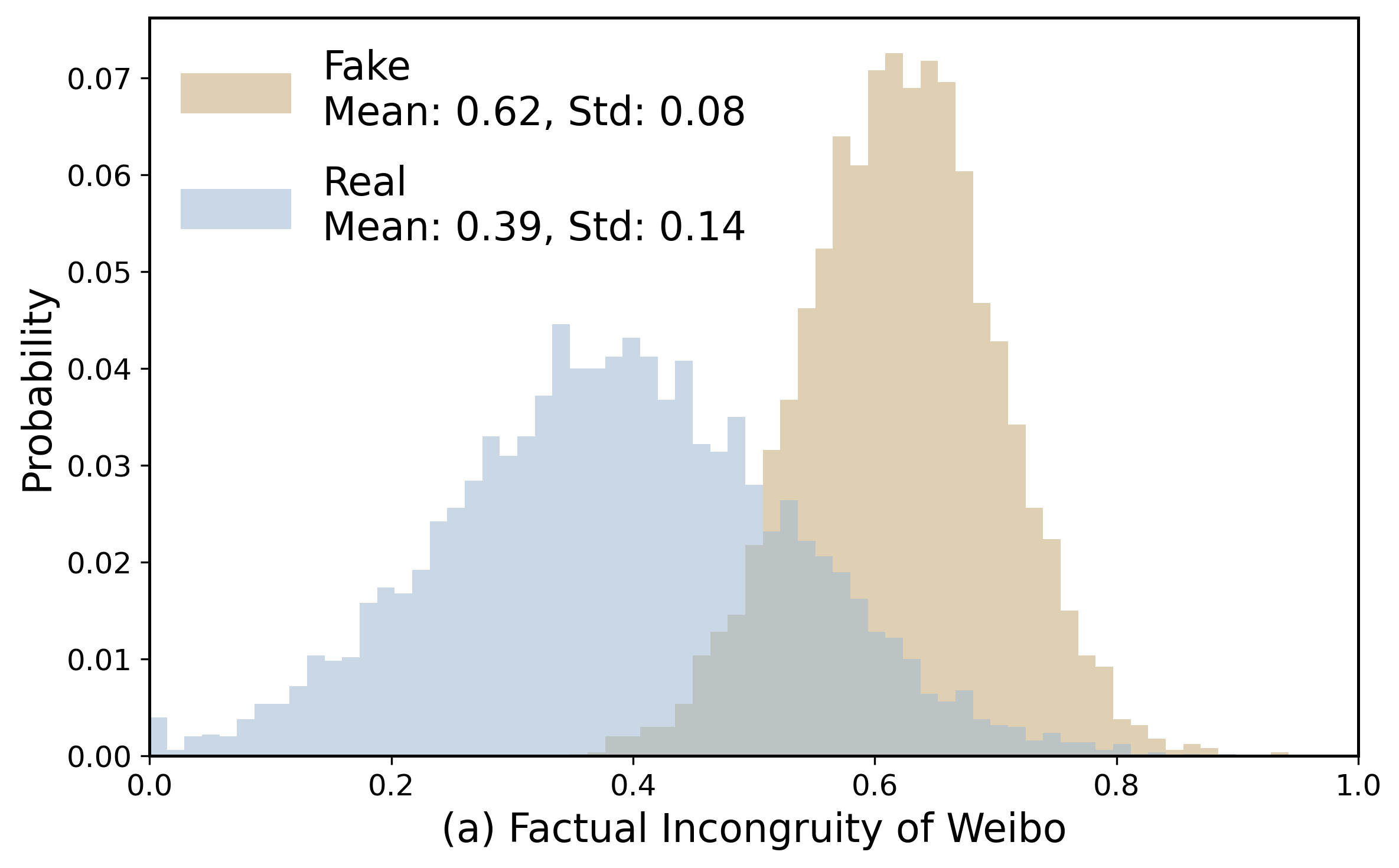}
        \label{fig1a}
    \end{subfigure}
    \begin{subfigure}{0.45\columnwidth}
        \includegraphics[width=\textwidth]{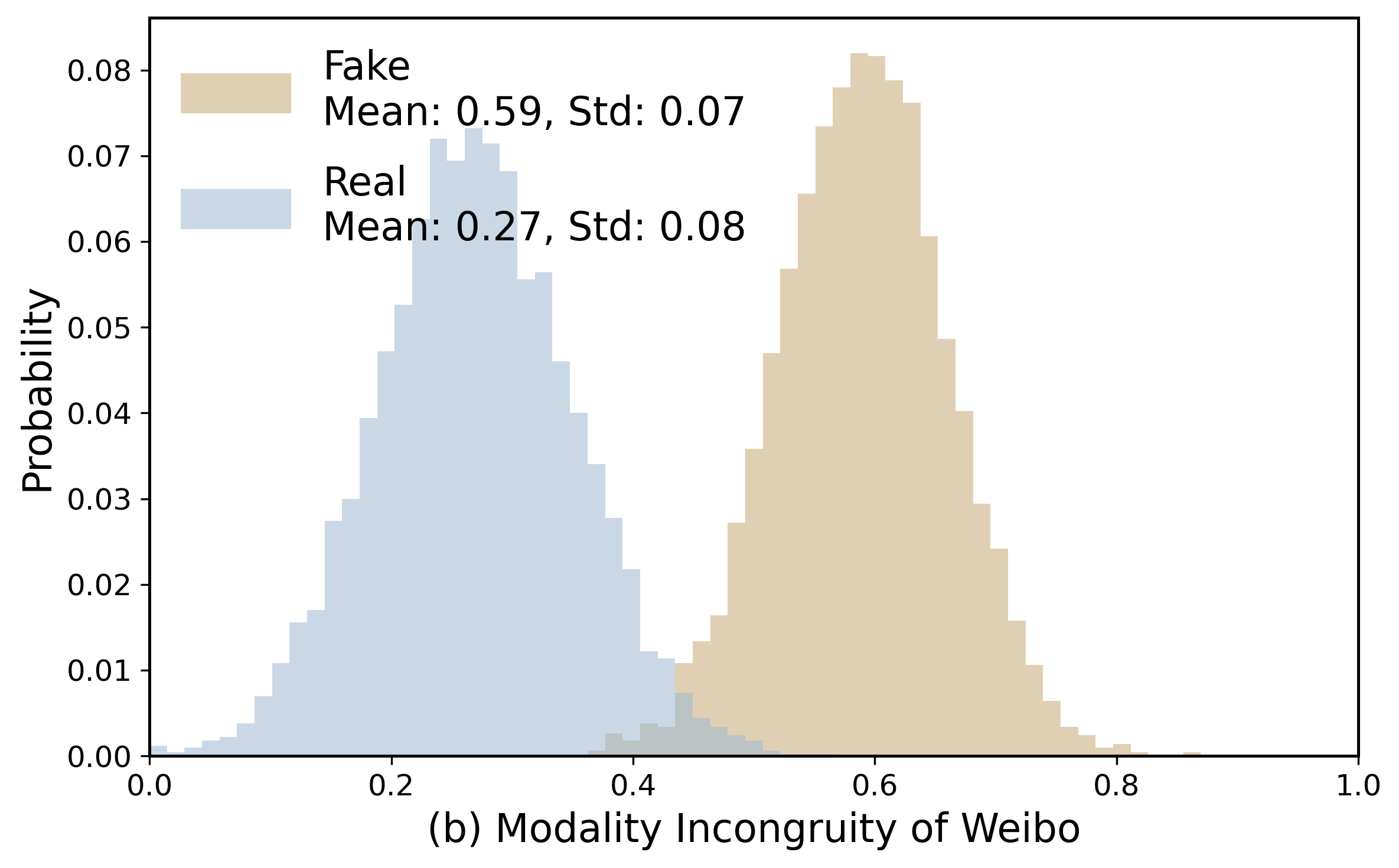}
        \label{fig1b}
    \end{subfigure}
    \begin{subfigure}{0.45\columnwidth}
        \includegraphics[width=\textwidth]{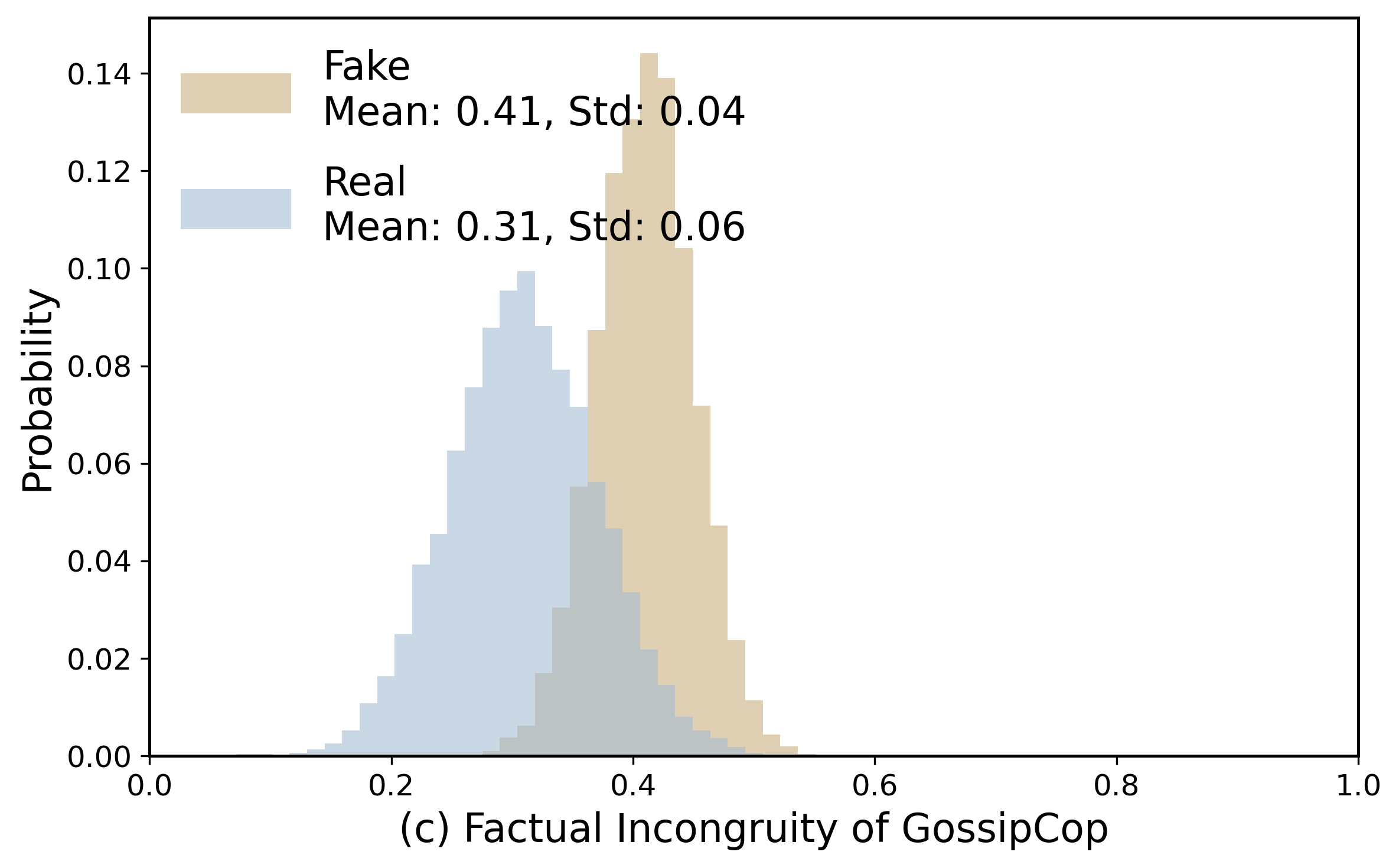}
        \label{fig1a}
    \end{subfigure}
    \begin{subfigure}{0.45\columnwidth}
        \includegraphics[width=\textwidth]{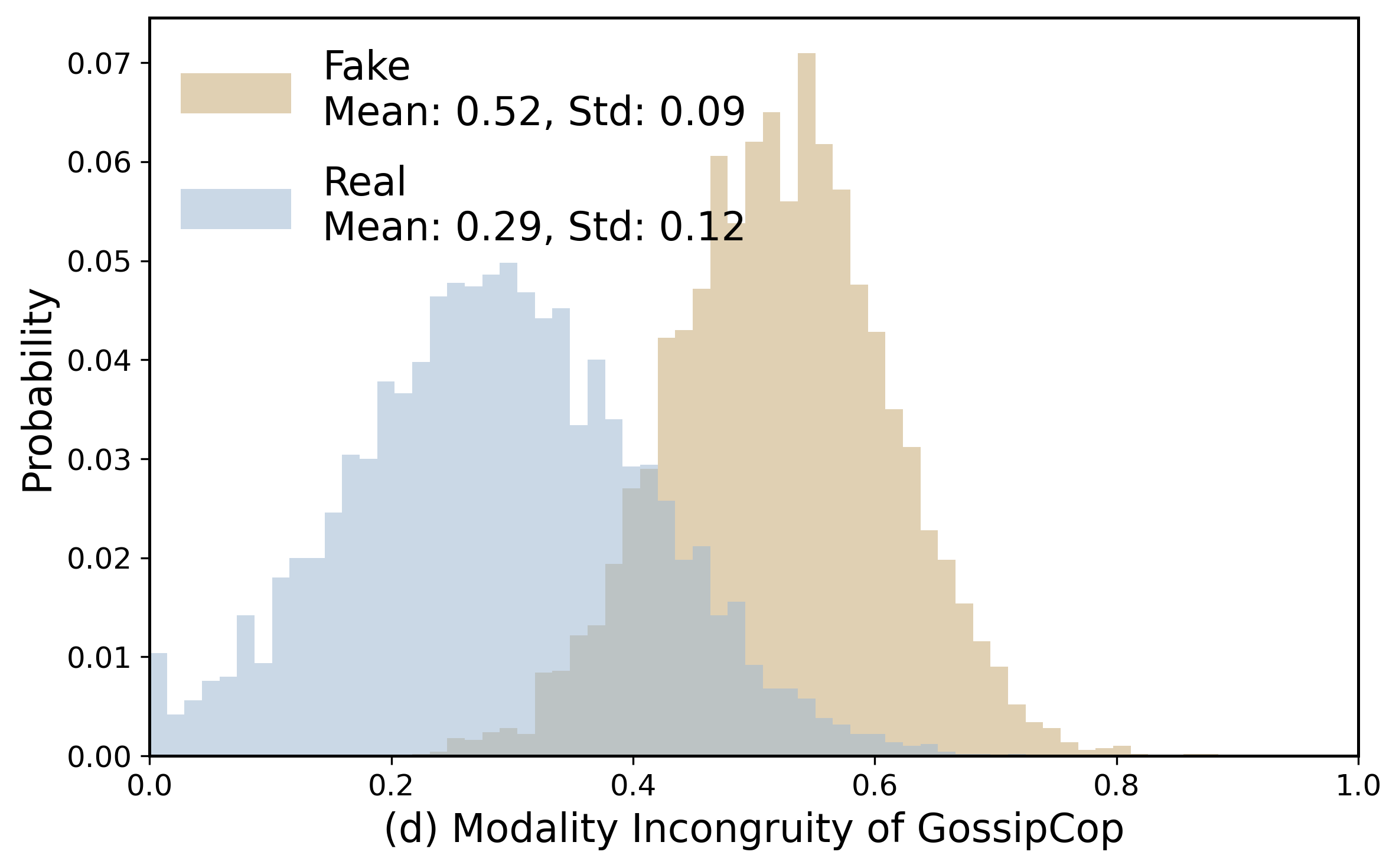}
        \label{fig1a}
    \end{subfigure}
    \caption{Distribution of incongruity from the fake news dataset. Left: Sample distribution exhibiting factual incongruity. Right: Sample distribution demonstrating modality incongruity. }
    \label{fig1}
\end{figure}

We explore two challenging incongruity issues in MFND tasks: factual and modality incongruity. As shown in Figure \ref{fig1}, the semantic similarity $s$ between modalities is obtained using CLIP \citep{CLIP}, and the factual incongruity is expressed as \(1 - s\). The BERT \citep{bert} and ViT \citep{vit} models are trained on textual and visual modalities, respectively, to quantify incongruity by the difference in predicted probabilities associated with news labels. The incongruity of fake news is significantly larger than that of real news at the modality level, especially as evidenced by the mean values, with a considerable number of samples having high modality incongruity. Meanwhile, this phenomenon also exists at the factual level. Inspired by the above research and validation, we designed our method to detect the incongruity of multi-modal fake news data at the modality and factual levels.

This paper proposes an Incongruity-aware Distribution Optimization (IDO), which consists of Factual Semantic  Distribution (FSD) modeling and Incongruity Contrastive  learning (ICL) modules. In FSD, based on semantic associations \citep{niu2021counterfactual,xu2025towards}, samples are distinguished by an adaptive strategy. Specifically, we maintain Gaussian distributions for fake and real samples, respectively, and use the probabilities they generate to model factual incongruity. Since the distribution depends on the extracted embeddings, we introduce a channel-based re-weighting strategy to learn a representation that judges news truthfulness. Modality incongruity is perceived by continuous contrastive learning. Overall, factual and modality information is reinforced in FSD and ICL and is used to model incongruity in MFND.

Our contributions are summarized as follows:
\begin{itemize}
    \item IDO explores and models incongruity in multimodal fake news detection.
    \item IDO learns to discriminate news authenticity from two tasks: factual and modality incongruity modeling. It leverages channel-wise reweighting and continuous contrast learning to model representations.
    \item Extensive experiments on public benchmarks demonstrate the effectiveness and superiority of IDO. 
\end{itemize}

\section{Methodology}
\subsection{Overview}
The image-text pair is formally defined as: \( I = \{ p^{i} \}_{i=1}^{m} \) and \( T = \{ w^{i} \}_{i=1}^{k} \), where \( p^{i} \) represents the \(i\)-th patch of the image, \( w^i \) is the \( i \)-th word. An image is split into \( m \) patches, and a text contains \( k \) words. The image and text are first processed by the image and text encoders, i.e. ViT \citep{vit}, BERT \citep{bert}. The output embeddings are defined as \( H_p \in \mathbb{R}^{m \times C} \) and \( H_w \in \mathbb{R}^{k \times C} \), \( C \) denotes the number of channels.

In order to combine the image and text modalities for judging the authenticity of news, we adopted a cross-modal attention module to implicitly construct the interaction between the image and text modalities. Specifically, with the embeddings \( H_p \in \mathbb{R}^{m \times C} \) and \( H_w \in \mathbb{R}^{k \times C} \), construct the relation matrix \( R \in \mathbb{R}^{m \times k} \) by matrix multiplication in the channel dimension, then pass the matrix through convolution layers:

\begin{equation}
\small
R = \text{Conv}(H_p \cdot H_w^{T}),
\end{equation}
where \(\text{Conv}\) is implemented by two convolution layers. The larger the value in \( R \) indicates a stronger correlation between different modalities. For the visual modality, we add the values of the textual tokens to generate the attention vector \( V_p \in \mathbb{R}^m \). The same way is adopted to form the attention vector \( V_w \in \mathbb{R}^k \) for the words. Then, \( V_p \) and \( V_w \) are integrated into \( H_p \) and \( H_w \) by channel-wise multiplication followed by the sigmoid activation. The aligned visual patch embeddings can be formalized as
\begin{equation}
\small
H_{ap} = \text{sigmoid}(V_p) \cdot H_p,
\end{equation}

Text embedding \( H_{aw} \) is processed in the same way. The explicit alignment enforces the image and text representations to be consistent by loss \citep{baltruvsaitis2018multimodal,radford2021learning}. However, multimodal fake news detection depends on the incongruity. The strong constraint harms the latent incongruity within the representations.  Then, the aligned embeddings are sent to FSD and ICL module.

\begin{table*}[ht]
\small
\centering

\begin{tabular}{clccccccc}
\hline
\multirow{2}{*}{\textbf{Dataset}} &
  \multicolumn{1}{c}{\multirow{2}{*}{\textbf{Method}}} &
  \multirow{2}{*}{\textbf{Accuracy}} &
  \multicolumn{3}{c}{\textbf{Fake News}} &
  \multicolumn{3}{c}{\textbf{Real News}} \\ \cline{4-9} 
\textbf{} &
  \multicolumn{1}{c}{\textbf{}} &
  \textbf{} &
  \textbf{Pre.} &
  \textbf{Rec.} &
  \textbf{F1} &
  \textbf{Pre.} &
  \textbf{Rec.} &
  \textbf{F1} \\ \hline
\multirow{10}{*}{Weibo}     & EANN \citep{EANN}     & 0.827 & 0.847 & 0.812 & 0.829 & 0.807 & 0.843 & 0.825 \\
                           & SpotFake \citep{SpotFake}  & 0.892 & 0.902 & \underline{0.964} & \underline{0.932} & 0.847 & 0.656 & 0.739 \\
                           & SAFE \citep{SAFE}      & 0.762 & 0.831 & 0.724 & 0.774 & 0.695 & 0.811 & 0.748 \\
                           & CAFE \citep{CAFE}       & 0.840 & 0.855 & 0.830 & 0.842 & 0.825 & 0.851 & 0.837 \\
                           & BMR \citep{BMR}       & 0.918 & 0.882 & 0.948 & 0.914 & \underline{0.942} & 0.879 & 0.904 \\
                           & FND-CLIP \citep{FNDCLIP}  & 0.907 & 0.914 & 0.901 & 0.908 & 0.914 & 0.901 & 0.907 \\
                           & MSACA \citep{MSACA} & 0.903 & \underline{0.935} & 0.873 & 0.903 & 0.872 & \underline{0.935} & 0.902 \\ 
                           & RaCMC \citep{RaCMC} & 0.915 & 0.910 & 0.924 & 0.917 & 0.921 & 0.906 & 0.914 \\ 
                           & MIMoE-FND \citep{MIMOEFND} & \underline{0.928} & \textbf{0.942} & 0.913 & 0.928 & 0.913 & \textbf{0.942} & \underline{0.927} \\
                           & \textbf{IDO}      & \textbf{0.947}     & 0.929     & \textbf{0.970}     & \textbf{0.949}     & \textbf{0.969}     & 0.924     & \textbf{0.946}     \\ \hline
\multirow{10}{*}{GossipCop} & EANN \citep{EANN}      & 0.864 & 0.702 & 0.518 & 0.594 & 0.887 & 0.956 & 0.920 \\
                           & SpotFake \citep{SpotFake}  & 0.858 & 0.732 & 0.372 & 0.494 & 0.866 & 0.962 & 0.914 \\
                           & SAFE \citep{SAFE}      & 0.838 & 0.758 & 0.558 & 0.643 & 0.857 & 0.937 & 0.895 \\
                           & CAFE \citep{CAFE}       & 0.867 & 0.732 & 0.490 & 0.587 & 0.887 & 0.957 & 0.921 \\
                           & BMR \citep{BMR}       & \underline{0.895} & 0.752 & 0.639 & 0.691 & \underline{0.920} & 0.965 & \underline{0.936} \\
                           & FND-CLIP \citep{FNDCLIP}  & 0.880 & 0.761 & 0.549 & 0.638 & 0.899 & 0.959 & 0.928 \\
                           & MSACA \citep{MSACA} & 0.887 & \underline{0.816} & 0.538 & 0.646 & 0.897 & \textbf{0.971} & 0.933 \\ 
                           & RaCMC \citep{RaCMC} & 0.879 & 0.745 & 0.563 & 0.641 & 0.902 & 0.954 & 0.927 \\ 
                           & MIMoE-FND \citep{MIMOEFND} & \underline{0.895} & 0.762 & \underline{0.644} & \underline{0.698}
                           & \underline{0.920} & 0.953 & \underline{0.936} \\
                           & \textbf{IDO}      & \textbf{0.912}     & \textbf{0.844}     & \textbf{0.695}     & \textbf{0.762}     & \textbf{0.926}     & \underline{0.967}     & \textbf{0.946}     \\ \hline
\multirow{8}{*}{Weibo-21}  & EANN \citep{EANN}      & 0.870 & 0.902 & 0.825 & 0.862 & 0.841 & 0.912 & 0.875 \\
                           & SpotFake \citep{SpotFake}  & 0.851 & \underline{0.953} & 0.733 & 0.828 & 0.786 & \textbf{0.964} & 0.866 \\
                           & SAFE \citep{SAFE}      & 0.905 & 0.893 & 0.908 & 0.901 & 0.916 & 0.901 & 0.890 \\
                           & CAFE \citep{CAFE}       & 0.882 & 0.857 & 0.915 & 0.885 & 0.907 & 0.844 & 0.876 \\
                           & BMR \citep{BMR}       & 0.929 & 0.908 & 0.947 & 0.927 & 0.946 & 0.906 & 0.925 \\
                           & FND-CLIP \citep{FNDCLIP} & 0.943 & 0.935 & 0.945 & 0.940 & 0.950 & 0.942 & 0.946 \\
                           & MIMoE-FND \citep{MIMOEFND} & \underline{0.956} & \underline{0.953} & \underline{0.957} & \underline{0.955} & \underline{0.959} & 0.956 & \underline{0.957} \\
                           & \textbf{IDO}      & \textbf{0.963}    & \textbf{0.960}     & \textbf{0.968}     & \textbf{0.961}     & \textbf{0.965}     & \underline{0.957}     & \textbf{0.961}     \\ \hline
\end{tabular}

\caption{Performance comparison with SOTA methods on MFND datasets. The best results are displayed in boldface, and the second-best results are displayed underlined.} 
\label{tab:main results} 
\end{table*}

\subsection{Factual Semantic Distribution Modeling}

The facts is perceived through semantic information \citep{niu2021counterfactual,yuan2025enhancing}. Factual Semantic Distribution (FSD) aims to intensify invariant representations that lead to misinformation, we utilize gaussian distributions to model the incongruity in the multi-modal data.

First, we introduce a channel-wise reweighting strategy to learn invariant representations. This strategy was formulated based on the observation that certain images are related to the content described by the text and do not convey misinformation. Inspired by the research about invariant risk minimization \citep{pmlr-v162-zhou22d}, we utilize reweighting to find the content most related to misinformation. Specifically, with the training of the model, representations related to misinformation are gradually activated. 

\begin{equation}
\small
R_p = H_{ap} \cdot \sigma(\text{ReLU}(FC(H_{ap}))),
\end{equation}

where \( R_p \) denotes reweighted embeddings for patches and \( \sigma \) means the channel-wise variance. The reweighted embeddings for words \( R_w \) is processed in the same way.

After acquiring the discriminative semantic embeddings, we maintain the similarity distributions of fake and real samples, and calculate the probability of multi-modal data. Specifically, for the \( R_p \) and \( R_w \), we utilize \( R_v \in \mathbb{R}^C \) and \( R_t \in \mathbb{R}^C \) as [CLS], which are calculated as the average of all the patch and word embeddings. During the training process, we maintain two memory banks \( M_{Fake} = \{(R_v^i, R_t^i)\}_{i=1}^q \), \( M_{Real} = \{(R_v^i, R_t^i)\}_{i=1}^q \) of fake and real semantic representations from previous batch, \( q \) represents the length of the memory bank. Based on the observation in Figure \ref{fig1}, we adopt the gaussian distribution, which can be estimated by the following formulas:

\begin{equation}
\small
\mu = \sum_{i=1}^q \text{Cos}(R_v^i, R_t^i),\sigma = \sqrt{\sum_{i=1}^q (\text{Cos}(R_v^i, R_t^i) - \mu)^2},
\end{equation}


where Cos denotes the cosine similarity function, \( \mu \) and \( \sigma \) are the mean and variance values of the maintained gaussian distribution. The distributions \( D_{Fake} \) and \( D_{Real} \) are denoted as \( D_{Fake} \in \mathcal{N}(\mu_{Fake}, \sigma_{Fake}) \), \( D_{Real} \in \mathcal{N}(\mu_{Real}, \sigma_{Real}) \). We model the possibility of the sample belonging to \( D_{Fake} \) and \( D_{Real} \) based on the probability density function.

\begin{equation}
\small
p = \frac{1}{\sigma \sqrt{2\pi}} \cdot e^{-\tau \left( \frac{\text{Cos}(H_v^i, H_t^i) - \mu}{\sigma} \right)^2},
\end{equation}

where \( \tau \) is the temperature controls the importance of \( \sigma \). The factual incongruity \( \lambda_{FSD} \) is calculated as \( p_{Real} - p_{Fake} \). Comparing with adopting the fixed or adaptive threshold to distinguish the misinformation, the Gaussian distribution provides a relatively gentle probability value, thereby avoiding the deviation caused by hard decision-making.

\subsection{Incongruity Contrastive Learning}
To model modality incongruity and  further boost the multi-modal discrimination representations, we introduce a continuous contrastive learning strategy, which constructs continuous supervision labels to capture the incongruity between modalities. The same as FSD, we use the average of the patch and word embeddings to obtain visual and textual [CLS]. Then, MLP was applied separately to obtain the representations \( p_t \in \mathbb{R}^B \) and \( p_v \in \mathbb{R}^B \)  of the text and the image for contrastive learning, where \( B \) is the mini-batch size. For an image-text pair, the large difference means the embeddings should be accordingly pushed away. Otherwise, they should be pulled close. Therefore, we construct the supervision \( G_p \) as follows:

\begin{equation}
\small
G_p^{ij} = \text{softmax}(\exp(-|p_v^i - p_t^j|)),
\end{equation}

The similarity matrix \( G_e \) of embeddings can be calculated by the dot product between the embeddings \( R_v \in \mathbb{R}^{B \times C} \) and \( R_t \in \mathbb{R}^{B \times C} \), which are the outputs of projection head:

\begin{equation}
\small
G_e^{ij} = \text{softmax}(\exp((R_v^i \cdot R_t^j))).
\end{equation}

Besides, the loss of continuous contrastive learning is calculated by the Kullback-Leibler (KL) divergence:

\begin{equation}
\small
\mathcal{L}_{ICL} = KL(G_e, G_p).
\end{equation}

Then, we model the incongruity between vision and text for fake news detection, denoted as \( \lambda_{ICL} = |p_v - p_t| \).

Considering the factors of factual inconsistency \( \lambda_{FSD} \) and modality inconsistency \( \lambda_{ICL} \),
\begin{equation}
\hat{y} = \operatorname{sigmoid}(\lambda_{\text{FSD}} + \lambda_{\text{ICL}}),
\end{equation}

The cross-entropy loss function used as:

\begin{equation}
\small
L_{\mathrm{CE}}(\boldsymbol{X}, \boldsymbol{Y})=-\frac{1}{n} \sum_{i=1}^{n} \boldsymbol{y}_{i}^{\top} \log \hat{\boldsymbol{y}}_{i}.
\end{equation}

Finally, the IDO network for multimodal fake news detection is optimized by the loss:

\begin{equation}
\small
\mathcal{L}_T = \mathcal{L}_{CE} + \alpha \mathcal{L}_{ICL}.
\end{equation}

 Where $\alpha$ is 0.5.

\section{Experimental Results}
\subsection{Main Results}

Table \ref{tab:main results} reports the performance comparison between our method and other baselines in detail.  IDO achieved state-of-the-art performance across all datasets, with accuracies of 0.947, 0.912 and 0.963 for Weibo, GossipCop and Weibo-21, respectively.  For Weibo-21, the performance of MIMoE-FND is quite comparable to that of our method. However, IDO significantly outperforms MIMoE-FND on Weibo and GossipCop datasets.

\begin{table}[t]
\centering
\small 
\begin{tabular}{clccc}
\hline
\textbf{Dataset}           & \textbf{Method} & \textbf{ACC} & \textbf{F1(Fake)} & \textbf{F1(Real)} \\ \hline
\multirow{5}{*}{Weibo}     & \textbf{IDO}            & \textbf{0.947}        & \textbf{0.949}             & \textbf{0.946}             \\
                           & w/o $\mathcal{L}_{ICL}$        & 0.930        & 0.927             & 0.932             \\
                           & w/o $\lambda_{ICL}$        & 0.931        & 0.931             & 0.932             \\
                           & w/o $\lambda_{FSD}$             & 0.941            & 0.942                 & 0.940                 \\
                           & Base             & 0.892            & 0.897                 & 0.887                 \\ \hline
\multirow{5}{*}{GossipCop} & \textbf{IDO}            & \textbf{0.912}        & \textbf{0.762}             & \textbf{0.946}             \\
                           & w/o $\mathcal{L}_{ICL}$        & 0.898        & 0.710             & 0.938             \\
                           & w/o  $\lambda_{ICL}$       & 0.895        & 0.699             & 0.936             \\
                           & w/o $\lambda_{FSD}$              & 0.907        & 0.743             & 0.943                 \\
                           & Base              & 0.871            & 0.663                 & 0.920                 \\ \hline
\multirow{5}{*}{Weibo21}   & \textbf{IDO}            & \textbf{0.963}       & \textbf{0.961}             & \textbf{0.961}             \\
                           & w/o $\mathcal{L}_{ICL}$        & 0.946        & 0.947             & 0.945             \\
                           & w/o $\lambda_{ICL}$        & 0.950        & 0.952             & 0.948             \\
                           & w/o $\lambda_{FSD}$             & 0.952            & 0.953                 & 0.951                 \\
                           & Base              & 0.922            & 0.925                 & 0.918                 \\ \hline
\end{tabular}
\caption{Ablation study on IDO. w/o stands for without. } 
\label{tab: Ablation study}
\end{table}


\subsection{Ablation Study}
To explore the effectiveness of each component in IDO for modeling incongruity, we carried out ablation experiments as shown in Table \ref{tab: Ablation study}. The results suggest four key observations. First, both FSD and ICL outperformed the baseline model. Second, using FSD to model incongruity improved accuracy, which shows the effectiveness of discriminative embedding. Third, ICL significantly boosted the F1(fake) value, proving the effectiveness of persistent contrastive learning in detecting fake news. Fourth, the model combining FSD and ICL performed the best, indicating the complementarity between the components.

\subsection{Optimal Settings Exploring}
\begin{figure}[t]
\centering
\includegraphics[width=0.4\textwidth]{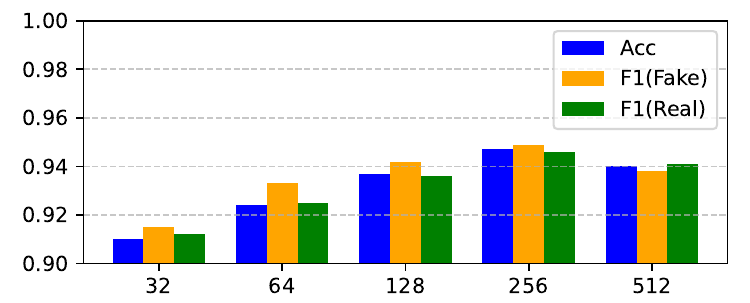} 
\caption{Analysis of memory bank capacity $L$.}
\label{fig-bank}
\end{figure}

In this section, we conduct experiments to determine the optimal settings for the memory bank of IDO. As shown in Figure \ref{fig-bank}, increasing the capacity ($L$) of the memory bank results in continuous improvements in accuracy and F1 score. However, when $L$ exceeds 256, the accuracy begins to decline. Therefore, a capacity size of 256 is selected as the optimal choice for modeling feature distributions, as it achieves the best performance. Conversely, increasing $L$ beyond this threshold degrades performance. This phenomenon may be attributed to an excessively large memory bank with a superabundance of instances makes it difficult to model the incongruity in feature distributions.

\section{Conclusion}
This paper presents IDO, a novel incongruity-aware distribution optimization framework for multimodal fake news detection. Experiments across three public MFND benchmarks demonstrate IDO’s superior performance over previous methods.




\section*{Impact Statement}

This paper presents a multimodal fake news detection method that aims to improve the identification of misleading content in online environments. The potential positive impact of this work is to support misinformation analysis and help reduce the spread of deceptive multimodal content.

However, such detection systems may also introduce risks, including false positives, false negatives, and potential biases inherited from datasets or model design. These issues may affect fairness, transparency, and reliability in real-world deployment. Therefore, this work should be used as an assistive tool rather than a substitute for human judgment, and future research should further investigate responsible deployment, robustness, and bias mitigation.

\bibliography{example_paper}

@article{10.1145/3395046,
author = {Zhou, Xinyi and Zafarani, Reza},
title = {A Survey of Fake News: Fundamental Theories, Detection Methods, and Opportunities},
year = {2020},
issue_date = {September 2021},
publisher = {Association for Computing Machinery},
address = {New York, NY, USA},
volume = {53},
number = {5},
issn = {0360-0300},
url = {https://doi.org/10.1145/3395046},
doi = {10.1145/3395046},
journal = {ACM Comput. Surv.},
month = sep,
articleno = {109},
numpages = {40},
}

@inproceedings{MVAE,
author = {Khattar, Dhruv and Goud, Jaipal Singh and Gupta, Manish and Varma, Vasudeva},
title = {MVAE: Multimodal Variational Autoencoder for Fake News Detection},
year = {2019},
isbn = {9781450366748},
publisher = {Association for Computing Machinery},
address = {New York, NY, USA},
url = {https://doi.org/10.1145/3308558.3313552},
doi = {10.1145/3308558.3313552},
booktitle = {The World Wide Web Conference},
pages = {2915–2921},
numpages = {7},
location = {San Francisco, CA, USA},
series = {WWW '19}
}

@INPROCEEDINGS{SpotFake,
  author={Singhal, Shivangi and Shah, Rajiv Ratn and Chakraborty, Tanmoy and Kumaraguru, Ponnurangam and Satoh, Shin'ichi},
  booktitle={2019 IEEE Fifth International Conference on Multimedia Big Data (BigMM)}, 
  title={SpotFake: A Multi-modal Framework for Fake News Detection}, 
  year={2019},
  volume={},
  number={},
  pages={39-47},
  doi={10.1109/BigMM.2019.00-44}}

@inproceedings{SAFE,
  title={SAFE: Similarity-Aware Multi-modal Fake News Detection},
  author={Zhou, Xinyi and Wu, Jindi and Zafarani, Reza},
  booktitle={Pacific-Asia Conference on Knowledge Discovery and Data Mining},
  pages={354--367},
  year={2020},
  organization={Springer}
}

@inproceedings{MCAN,
    title = "Multimodal Fusion with Co-Attention Networks for Fake News Detection",
    author = "Wu, Yang  and
      Zhan, Pengwei  and
      Zhang, Yunjian  and
      Wang, Liming  and
      Xu, Zhen",
    editor = "Zong, Chengqing  and
      Xia, Fei  and
      Li, Wenjie  and
      Navigli, Roberto",
    booktitle = "Findings of the Association for Computational Linguistics: ACL-IJCNLP 2021",
    month = aug,
    year = "2021",
    address = "Online",
    publisher = "Association for Computational Linguistics",
    url = "https://aclanthology.org/2021.findings-acl.226/",
    doi = "10.18653/v1/2021.findings-acl.226",
    pages = "2560--2569"
}

@article{MCNN,
title = {Detecting fake news by exploring the consistency of multimodal data},
journal = {Information Processing Management},
volume = {58},
number = {5},
pages = {102610},
year = {2021},
issn = {0306-4573},
doi = {https://doi.org/10.1016/j.ipm.2021.102610},
url = {https://www.sciencedirect.com/science/article/pii/S0306457321001060},
author = {Junxiao Xue and Yabo Wang and Yichen Tian and Yafei Li and Lei Shi and Lin Wei},
}

@inproceedings{CAFE,
author = {Chen, Yixuan and Li, Dongsheng and Zhang, Peng and Sui, Jie and Lv, Qin and Tun, Lu and Shang, Li},
title = {Cross-modal Ambiguity Learning for Multimodal Fake News Detection},
year = {2022},
isbn = {9781450390965},
publisher = {Association for Computing Machinery},
address = {New York, NY, USA},
url = {https://doi.org/10.1145/3485447.3511968},
doi = {10.1145/3485447.3511968},
booktitle = {Proceedings of the ACM Web Conference 2022},
pages = {2897–2905},
numpages = {9},
location = {Virtual Event, Lyon, France},
series = {WWW '22}
}

@inproceedings{COOLANT,
author = {Wang, Longzheng and Zhang, Chuang and Xu, Hongbo and Xu, Yongxiu and Xu, Xiaohan and Wang, Siqi},
title = {Cross-modal Contrastive Learning for Multimodal Fake News Detection},
year = {2023},
isbn = {9798400701085},
publisher = {Association for Computing Machinery},
address = {New York, NY, USA},
url = {https://doi.org/10.1145/3581783.3613850},
doi = {10.1145/3581783.3613850},
booktitle = {Proceedings of the 31st ACM International Conference on Multimedia},
pages = {5696–5704},
numpages = {9},
location = {Ottawa ON, Canada},
series = {MM '23}
}

@inproceedings{BMR,
author = {Ying, Qichao and Hu, Xiaoxiao and Zhou, Yangming and Qian, Zhenxing and Zeng, Dan and Ge, Shiming},
title = {Bootstrapping multi-view representations for fake news detection},
year = {2023},
isbn = {978-1-57735-880-0},
publisher = {AAAI Press},
url = {https://doi.org/10.1609/aaai.v37i4.25670},
doi = {10.1609/aaai.v37i4.25670},
booktitle = {Proceedings of the Thirty-Seventh AAAI Conference on Artificial Intelligence and Thirty-Fifth Conference on Innovative Applications of Artificial Intelligence and Thirteenth Symposium on Educational Advances in Artificial Intelligence},
articleno = {601},
numpages = {9},
series = {AAAI'23/IAAI'23/EAAI'23}
}

@INPROCEEDINGS{FNDCLIP,
  author={Zhou, Yangming and Yang, Yuzhou and Ying, Qichao and Qian, Zhenxing and Zhang, Xinpeng},
  booktitle={2023 IEEE International Conference on Multimedia and Expo (ICME)}, 
  title={Multimodal Fake News Detection via CLIP-Guided Learning}, 
  year={2023},
  volume={},
  number={},
  pages={2825-2830},
  doi={10.1109/ICME55011.2023.00480}}

@inproceedings{MIMOEFND,
author = {Liu, Yifan and Liu, Yaokun and Li, Zelin and Yao, Ruichen and Zhang, Yang and Wang, Dong},
title = {Modality Interactive Mixture-of-Experts for Fake News Detection},
year = {2025},
isbn = {9798400712746},
publisher = {Association for Computing Machinery},
address = {New York, NY, USA},
url = {https://doi.org/10.1145/3696410.3714522},
doi = {10.1145/3696410.3714522},
booktitle = {Proceedings of the ACM on Web Conference 2025},
pages = {5139–5150},
numpages = {12},
location = {Sydney NSW, Australia},
series = {WWW '25}
}

@inproceedings{EANN,
author = {Wang, Yaqing and Ma, Fenglong and Jin, Zhiwei and Yuan, Ye and Xun, Guangxu and Jha, Kishlay and Su, Lu and Gao, Jing},
title = {EANN: Event Adversarial Neural Networks for Multi-Modal Fake News Detection},
year = {2018},
isbn = {9781450355520},
publisher = {Association for Computing Machinery},
address = {New York, NY, USA},
url = {https://doi.org/10.1145/3219819.3219903},
doi = {10.1145/3219819.3219903},
booktitle = {Proceedings of the 24th ACM SIGKDD International Conference on Knowledge Discovery \& Data Mining},
pages = {849–857},
numpages = {9},
location = {London, United Kingdom},
series = {KDD '18}
}

@inproceedings{MSACA,
author = {Wang, Jiandong and Zhang, Hongguang and Liu, Chun and Yang, Xiongjun},
title = {Fake News Detection via Multi-scale Semantic Alignment and Cross-modal Attention},
year = {2024},
isbn = {9798400704314},
publisher = {Association for Computing Machinery},
address = {New York, NY, USA},
url = {https://doi.org/10.1145/3626772.3657905},
doi = {10.1145/3626772.3657905},
booktitle = {Proceedings of the 47th International ACM SIGIR Conference on Research and Development in Information Retrieval},
pages = {2406–2410},
numpages = {5},
location = {Washington DC, USA},
series = {SIGIR '24}
}

@article{RaCMC, title={RaCMC: Residual-Aware Compensation Network with Multi-Granularity Constraints for Fake News Detection}, volume={39}, url={https://ojs.aaai.org/index.php/AAAI/article/view/32084}, DOI={10.1609/aaai.v39i1.32084}, abstractNote={Multimodal fake news detection aims to automatically identify real or fake news, thereby mitigating the adverse effects caused by such misinformation. Although prevailing approaches have demonstrated their effectiveness, challenges persist in cross-modal feature fusion and refinement for classification. To address this, we present a residual-aware compensation network with multi-granularity constraints (RaCMC) for fake news detection, that aims to sufficiently interact and fuse cross-modal features while amplifying the differences between real and fake news. First, a multiscale residual-aware compensation module is designed to interact and fuse features at different scales, and ensure both the consistency and exclusivity of feature interaction, thus acquiring high-quality features. Second, a multi-granularity constraints module is implemented to limit the distribution of both the news overall and the image-text pairs within the news, thus amplifying the differences between real and fake news at the news and feature levels. Finally, a dominant feature fusion reasoning module is developed to comprehensively evaluate news authenticity from the perspectives of both consistency and inconsistency. Experiments on three public datasets, including Weibo17, Politifact and GossipCop, reveal the superiority of the proposed method.}, number={1}, journal={Proceedings of the AAAI Conference on Artificial Intelligence}, author={Yu, Xinquan and Sheng, Ziqi and Lu, Wei and Luo, Xiangyang and Zhou, Jiantao}, year={2025}, month={Apr.}, pages={986-994} }

@ARTICLE{weibo,
  author={Jin, Zhiwei and Cao, Juan and Zhang, Yongdong and Zhou, Jianshe and Tian, Qi},
  journal={IEEE Transactions on Multimedia}, 
  title={Novel Visual and Statistical Image Features for Microblogs News Verification}, 
  year={2017},
  volume={19},
  number={3},
  pages={598-608},
  doi={10.1109/TMM.2016.2617078}}

@inproceedings{weibo21,
author = {Nan, Qiong and Cao, Juan and Zhu, Yongchun and Wang, Yanyan and Li, Jintao},
title = {MDFEND: Multi-domain Fake News Detection},
year = {2021},
isbn = {9781450384469},
publisher = {Association for Computing Machinery},
address = {New York, NY, USA},
url = {https://doi.org/10.1145/3459637.3482139},
doi = {10.1145/3459637.3482139},
booktitle = {Proceedings of the 30th ACM International Conference on Information \& Knowledge Management},
pages = {3343–3347},
numpages = {5},
location = {Virtual Event, Queensland, Australia},
series = {CIKM '21}
}

@article{shu2018fakenewsnet,
  title={FakeNewsNet: A Data Repository with News Content, Social Context and Dynamic Information for Studying Fake News on Social Media},
  author={Shu, Kai and  Mahudeswaran, Deepak and Wang, Suhang and Lee, Dongwon and Liu, Huan},
  journal={arXiv preprint arXiv:1809.01286},
  year={2018}
}

@inproceedings{wu2022characterizing,
  title={Characterizing and overcoming the greedy nature of learning in multi-modal deep neural networks},
  author={Wu, Nan and Jastrzebski, Stanislaw and Cho, Kyunghyun and Geras, Krzysztof J},
  booktitle={International Conference on Machine Learning},
  pages={24043--24055},
  year={2022},
  organization={PMLR}
}

@inproceedings{peng2022balanced,
  title={Balanced multimodal learning via on-the-fly gradient modulation},
  author={Peng, Xiaokang and Wei, Yake and Deng, Andong and Wang, Dong and Hu, Di},
  booktitle={Proceedings of the IEEE/CVF conference on computer vision and pattern recognition},
  pages={8238--8247},
  year={2022}
}

@inproceedings{fan2023pmr,
  title={Pmr: Prototypical modal rebalance for multimodal learning},
  author={Fan, Yunfeng and Xu, Wenchao and Wang, Haozhao and Wang, Junxiao and Guo, Song},
  booktitle={Proceedings of the IEEE/CVF Conference on Computer Vision and Pattern Recognition},
  pages={20029--20038},
  year={2023}
}

@inproceedings{li2023boosting,
  title={Boosting multi-modal model performance with adaptive gradient modulation},
  author={Li, Hong and Li, Xingyu and Hu, Pengbo and Lei, Yinuo and Li, Chunxiao and Zhou, Yi},
  booktitle={Proceedings of the IEEE/CVF International Conference on Computer Vision},
  pages={22214--22224},
  year={2023}
}

@inproceedings{hua2024reconboost,
  title={ReconBoost: Boosting Can Achieve Modality Reconcilement},
  author={Hua, Cong and Xu, Qianqian and Bao, Shilong and Yang, Zhiyong and Huang, Qingming},
  booktitle={International Conference on Machine Learning},
  pages={19573--19597},
  year={2024},
  organization={PMLR}
}

@inproceedings{wei2024mmpareto,
  title={MMPareto: Boosting Multimodal Learning with Innocent Unimodal Assistance},
  author={Wei, Yake and Hu, Di},
  booktitle={International Conference on Machine Learning},
  pages={52559--52572},
  year={2024},
  organization={PMLR}
}

@article{yang2024facilitating,
  title={Facilitating multimodal classification via dynamically learning modality gap},
  author={Yang, Yang and Wan, Fengqiang and Jiang, Qing-Yuan and Xu, Yi},
  journal={Advances in Neural Information Processing Systems},
  volume={37},
  pages={62108--62122},
  year={2024}
}

@article{guo2024classifier,
  title={Classifier-guided gradient modulation for enhanced multimodal learning},
  author={Guo, Zirun and Jin, Tao and Chen, Jingyuan and Zhao, Zhou},
  journal={Advances in Neural Information Processing Systems},
  volume={37},
  pages={133328--133344},
  year={2024}
}

@inproceedings{CLIP,
  title={Learning transferable visual models from natural language supervision},
  author={Radford, Alec and Kim, Jong Wook and Hallacy, Chris and Ramesh, Aditya and Goh, Gabriel and Agarwal, Sandhini and Sastry, Girish and Askell, Amanda and Mishkin, Pamela and Clark, Jack and others},
  booktitle={International conference on machine learning},
  pages={8748--8763},
  year={2021},
  organization={PmLR}
}

@article{vit,
  title={An image is worth 16x16 words: Transformers for image recognition at scale},
  author={Dosovitskiy, Alexey and Beyer, Lucas and Kolesnikov, Alexander and Weissenborn, Dirk and Zhai, Xiaohua and Unterthiner, Thomas and Dehghani, Mostafa and Minderer, Matthias and Heigold, Georg and Gelly, Sylvain and others},
  journal={arXiv preprint arXiv:2010.11929},
  year={2020}
}

@inproceedings{bert,
  title={Bert: Pre-training of deep bidirectional transformers for language understanding},
  author={Devlin, Jacob and Chang, Ming-Wei and Lee, Kenton and Toutanova, Kristina},
  booktitle={Proceedings of the 2019 conference of the North American chapter of the association for computational linguistics: human language technologies, volume 1 (long and short papers)},
  pages={4171--4186},
  year={2019}
}

@article{goldstein2023generative,
  title={Generative language models and automated influence operations: Emerging threats and potential mitigations},
  author={Goldstein, Josh A and Sastry, Girish and Musser, Micah and DiResta, Renee and Gentzel, Matthew and Sedova, Katerina},
  journal={arXiv preprint arXiv:2301.04246},
  year={2023}
}

@inproceedings{nslm,
  title={Unveiling implicit deceptive patterns in multi-modal fake news via neuro-symbolic reasoning},
  author={Dong, Yiqi and He, Dongxiao and Wang, Xiaobao and Jin, Youzhu and Ge, Meng and Yang, Carl and Jin, Di},
  booktitle={Proceedings of the AAAI Conference on Artificial Intelligence},
  volume={38},
  number={8},
  pages={8354--8362},
  year={2024}
}

@inproceedings{fka-owl,
  title={Fka-owl: Advancing multimodal fake news detection through knowledge-augmented lvlms},
  author={Liu, Xuannan and Li, Peipei and Huang, Huaibo and Li, Zekun and Cui, Xing and Liang, Jiahao and Qin, Lixiong and Deng, Weihong and He, Zhaofeng},
  booktitle={Proceedings of the 32nd ACM International Conference on Multimedia},
  pages={10154--10163},
  year={2024}
}

@article{cao2020exploring,
  title={Exploring the role of visual content in fake news detection},
  author={Cao, Juan and Qi, Peng and Sheng, Qiang and Yang, Tianyun and Guo, Junbo and Li, Jintao},
  journal={Disinformation, Misinformation, and Fake News in Social Media: Emerging Research Challenges and Opportunities},
  pages={141--161},
  year={2020},
  publisher={Springer}
}

@article{zhou2025robustrealiblemultimodalmisinformation,
  title={Towards Robust and Relible Multimodal Misinformation Recognition with Incomplete Modality},
  author={Zhou, Hengyang and Wei, Yiwei and Yang, Jian and Zhang, Zhenyu},
  journal={arXiv preprint arXiv:2510.05839},
  year={2025}
}

@InProceedings{pmlr-v162-zhou22d,
  title = 	 {Model Agnostic Sample Reweighting for Out-of-Distribution Learning},
  author =       {Zhou, Xiao and Lin, Yong and Pi, Renjie and Zhang, Weizhong and Xu, Renzhe and Cui, Peng and Zhang, Tong},
  booktitle = 	 {Proceedings of the 39th International Conference on Machine Learning},
  pages = 	 {27203--27221},
  year = 	 {2022},
  editor = 	 {Chaudhuri, Kamalika and Jegelka, Stefanie and Song, Le and Szepesvari, Csaba and Niu, Gang and Sabato, Sivan},
  volume = 	 {162},
  series = 	 {Proceedings of Machine Learning Research},
  month = 	 {17--23 Jul},
  publisher =    {PMLR},
  url = 	 {https://proceedings.mlr.press/v162/zhou22d.html}
}

@article{baltruvsaitis2018multimodal,
  title={Multimodal machine learning: A survey and taxonomy},
  author={Baltru{\v{s}}aitis, Tadas and Ahuja, Chaitanya and Morency, Louis-Philippe},
  journal={IEEE transactions on pattern analysis and machine intelligence},
  volume={41},
  number={2},
  pages={423--443},
  year={2018},
  publisher={IEEE}
}

@inproceedings{radford2021learning,
  title={Learning transferable visual models from natural language supervision},
  author={Radford, Alec and Kim, Jong Wook and Hallacy, Chris and Ramesh, Aditya and Goh, Gabriel and Agarwal, Sandhini and Sastry, Girish and Askell, Amanda and Mishkin, Pamela and Clark, Jack and others},
  booktitle={International conference on machine learning},
  pages={8748--8763},
  year={2021},
  organization={PmLR}
}

@inproceedings{niu2021counterfactual,
  title={Counterfactual vqa: A cause-effect look at language bias},
  author={Niu, Yulei and Tang, Kaihua and Zhang, Hanwang and Lu, Zhiwu and Hua, Xian-Sheng and Wen, Ji-Rong},
  booktitle={Proceedings of the IEEE/CVF conference on computer vision and pattern recognition},
  pages={12700--12710},
  year={2021}
}

@inproceedings{wei2024g,
  title={G\^{} 2SAM: Graph-Based Global Semantic Awareness Method for Multimodal Sarcasm Detection},
  author={Wei, Yiwei and Yuan, Shaozu and Zhou, Hengyang and Wang, Longbiao and Yan, Zhiling and Yang, Ruosong and Chen, Meng},
  booktitle={Proceedings of the AAAI Conference on Artificial Intelligence},
  volume={38},
  number={8},
  pages={9151--9159},
  year={2024}
}

@article{wei2024towards,
  title={Towards multimodal sarcasm detection via label-aware graph contrastive learning with back-translation augmentation},
  author={Wei, Yiwei and Duan, Maomao and Zhou, Hengyang and Jia, Zhiyang and Gao, Zengwei and Wang, Longbiao},
  journal={Knowledge-Based Systems},
  volume={300},
  pages={112109},
  year={2024},
  publisher={Elsevier}
}

@inproceedings{zhou2025ldgnet,
  title={LDGNet: LLMs Debate-Guided Network for Multimodal Sarcasm Detection},
  author={Zhou, Hengyang and Yan, Jinwu and Chen, Yaqing and Hong, Rongman and Zuo, Wenbo and Jin, Keyan},
  booktitle={ICASSP 2025-2025 IEEE International Conference on Acoustics, Speech and Signal Processing (ICASSP)},
  pages={1--5},
  year={2025},
  organization={IEEE}
}

@article{wei2025deepmsd,
  title={DeepMSD: Advancing Multimodal Sarcasm Detection through Knowledge-augmented Graph Reasoning},
  author={Wei, Yiwei and Zhou, Hengyang and Yuan, Shaozu and Chen, Meng and Shi, Haitao and Jia, Zhiyang and Wang, Longbiao and He, Xiaodong},
  journal={IEEE Transactions on Circuits and Systems for Video Technology},
  year={2025},
  publisher={IEEE}
}

@misc{zhou2026diverdynamiciterativevisual,
      title={DIVER: Dynamic Iterative Visual Evidence Reasoning for Multimodal Fake News Detection}, 
      author={Weilin Zhou and Zonghao Ying and Chunlei Meng and Jiahui Liu and Hengyang Zhou and Quanchen Zou and Deyue Zhang and Dongdong Yang and Xiangzheng Zhang},
      year={2026},
      eprint={2601.07178},
      archivePrefix={arXiv},
      primaryClass={cs.CV},
      url={https://arxiv.org/abs/2601.07178}, 
}

@article{yuan2025enhancing,
  title={Enhancing Semantic Awareness by Sentimental Constraint with Automatic Outlier Masking for Multimodal Sarcasm Detection},
  author={Yuan, Shaozu and Wei, Yiwei and Zhou, Hengyang and Xu, Qinfu and Chen, Meng and He, Xiaodong},
  journal={IEEE Transactions on Multimedia},
  year={2025},
  publisher={IEEE}
}

@inproceedings{xu2025towards,
  title={Towards Multimodal Sentiment Analysis via Hierarchical Correlation Modeling with Semantic Distribution Constraints},
  author={Xu, Qinfu and Wei, Yiwei and Wu, Chunlei and Wang, Leiquan and Yuan, Shaozu and Wu, Jie and Lu, Jing and Zhou, Hengyang},
  booktitle={Proceedings of the AAAI Conference on Artificial Intelligence},
  volume={39},
  number={20},
  pages={21788--21796},
  year={2025}
}

@article{pan2026beyond,
  title={Beyond Isolated Utterances: Cue-Guided Interaction for Context-Dependent Conversational Multimodal Understanding},
  author={Pan, Zhaoyan and Zhou, Hengyang and Li, Xiangdong and Wang, Yuning and Lou, Ye and Pan, Jiatong and Zhou, Ji and Zhang, Wei},
  journal={arXiv preprint arXiv:2604.25618},
  year={2026}
}

@misc{pan2026stateanchoredcompleteviewdistillationrobust,
      title={State-Anchored Complete-View Distillation for Robust Conversational Multimodal Emotion Recognition}, 
      author={Zhaoyan Pan and Xiangdong Li and Wenke Wu and Mengting Ma and Ye Lou and Ji Zhou and Jiatong Pan and Wei Zhang},
      year={2026},
      eprint={2605.29590},
      archivePrefix={arXiv},
      primaryClass={cs.MM},
      url={https://arxiv.org/abs/2605.29590}, 
}
\bibliographystyle{icml2026}

\newpage
\appendix
\onecolumn
\section{Related Work}
\subsection{Multimodal Fake News Detection}
In recent years, multimodal fake news detection has emerged as a significant research task \citep{10.1145/3395046}. MVAE \citep{MVAE} uses variational auto-encoders to learn a shared text-image embedding space, improving multimodal fusion. SpotFake \citep{SpotFake} leverages pre-trained models for feature extraction, benefiting from the rich features provided by these large-scale models. SAFE \citep{SAFE} models cross-modal interaction by computing the similarity between textual and visual information. MCAN \citep{MCAN} proposes a co-attention mechanism for learning fused features from visual and textual modality. MCNN \citep{MCNN} introduces a weight-sharing scheme that models cross-modal consistency through the computation of cosine similarity between cross-modal representations. CAFE \citep{CAFE} utilizes probabilistic modeling with two VAEs for textual and visual distribution modeling. COOLANT \citep{COOLANT} incorporates cross-modal contrastive learning to address this task. BMR \citep{BMR} dynamically fuses multi-perspective predictions via an improved MMoE network. FND-CLIP \citep{FNDCLIP} introduces a framework that utilizes CLIP’s pre-trained features for multimodal fusion, capturing cross-modal consistency distributions across semantic hierarchies. \citep{MSACA} propose MSACA, aligning text-image modalities at multiple scales via hierarchical representations, semantic consistency enhancement, and cross-modal attention. RaCMC \citep{RaCMC} introduces residual-aware multiscale fusion and multi-granularity constraints to enhance real/fake feature separation and alignment. MMLNet \citep{zhou2025robustrealiblemultimodalmisinformation} is a method for detecting fake news in incomplete modalities. MIMoE-FND \citep{MIMOEFND} introduces modality interaction modeling and proposes a hierarchical MoE framework that adaptively routes instances to fusion experts based on learned interaction patterns. DIVER \citep{zhou2026diverdynamiciterativevisual} proposed a framework based on iterative reasoning using visual information. However, the previous approach focused mainly on modeling the overall consistency, without paying attention to the inconsistencies at the modal and factual levels.

\subsection{Multimodal Learning}
One of the primary challenges in multimodal learning is how to effectively leverage and integrate information from diverse modalities, typically utilizing the complementary information among modalities for downstream tasks. \citep{wu2022characterizing} proposed conditional learning rates to facilitate multimodal learning by balancing relative learning speeds.  OGMGE \citep{peng2022balanced} proposed a modality contribution-based gradient update strategy, allowing modalities which contribute more to the learning objective to dominate the gradient updates. PMR \citep{fan2023pmr} via a prototype modality rebalancing strategy to accelerate learning in slower-learning modalities without compromising other modalities. AGM \citep{li2023boosting} proposed an adaptive gradient modulation method to improve the performance of jointly trained multimodal models. ReconBoost \citep{hua2024reconboost} proposed a modality alternating learning paradigm to mediate competition between dominant and weak modality. MMPareto \citep{wei2024mmpareto} mitigates gradient conflicts between multimodal and unimodal learning objectives by ensuring the final gradient maintains a common direction applicable to all learning objectives. \citep{wei2024g,wei2024towards} proposed label-based contrastive learning, which is used to optimize the distribution of the feature space. \citep{wei2025deepmsd, zhou2025ldgnet} generates external knowledge through multimodal large models and enhances multimodal learning. \citep{yang2024facilitating} addressed modality imbalance by integrating unsupervised and supervised learning approaches. CGGM \citep{guo2024classifier} proposed a classifier-guided gradient modulation method that enhances multimodal learning through joint modulation of both gradient magnitude and direction. CUCI-Net \citep{pan2026beyond} uniquely abstracts context-utterance dependency into an explicit cue for multimodal interaction. CoRe-KD \citep{pan2026stateanchoredcompleteviewdistillationrobust} handles missing modalities via state anchoring and conflict regularization without input reconstruction in multimodal learning.


\section{Experimental Setup}

\subsection{Datasets}
We conducted experiments on three real-world, cross-lingual benchmarks to evaluate the model's effectiveness: Weibo \citep{weibo}, Weibo21 \citep{weibo21}, and GossipCop \citep{shu2018fakenewsnet}, which are among the most commonly used multimodal fake news detection datasets. The Weibo and Weibo21 datasets originate from the Chinese social media platform Weibo. The GossipCop dataset comprises entertainment news articles collected from the GossipCop website.  

\subsection{Comparison Methods}
To demonstrate the superiority of the proposed method in multimodal fake news detection, we compared it with 9 baseline models, including EANN \citep{EANN}, SAFE \citep{SAFE}, SpotFake \citep{SpotFake}, CAFE \citep{CAFE}, BMR \citep{BMR}, FND-CLIP \citep{FNDCLIP}, MSACA \citep{MSACA}, RaCMC \citep{RaCMC}, and MIMoE-FND \citep{MIMOEFND}.

\subsection{Implementation Details}

To ensure reproducibility, we conducted experiments with three different random seeds (42, 66, 88) and reported the averaged results. For the image, the size is first resized to 224 × 224 and then divided into 32×32 patches, ViT adopts clip-vit-base-patch32. We use bert-base-chinese for Chinese texts and bert-base-uncased  for English texts. The memory bank has a capacity of 256 elements. The batch size is set to 16. The network is optimized by stochastic gradient descent with weight decay of 1e-5. The model was trained for 20 epochs. The learning rate is set to 2e-5 for the image and text encoders and 5e-5 for the rest. The experiments are implemented on a Linux server equipped with two GeForce RTX4090 GPUs.


\end{document}